\documentclass[manuscript,screen]{acmart}

\usepackage{algorithm}	
\usepackage[noend]{algpseudocode}
\usepackage{graphicx} 
\usepackage{multicol}        
\usepackage[bottom]{footmisc}
\usepackage{subfig}

\AtBeginDocument{
  \providecommand\BibTeX{{
    \normalfont B\kern-0.5em{\scshape i\kern-0.25em b}\kern-0.8em\TeX}}}

\setcopyright{acmcopyright}
\copyrightyear{2018}
\acmYear{2018}
\acmDOI{10.1007/978-981-15-6844-2_11}

\acmConference[HPVI '20]{High-Performance Vision Intelligence}{Springer}{Singapore}

\acmBooktitle{High-Performance Vision Intelligence} 
\acmISBN{978-981-15-6844-2}

\begin{document}

\title{Semantic Image Completion and Enhancement using GANs}

\author{Priyansh Saxena}
\orcid{0000-0003-1407-9752}
\affiliation{
  \institution{ABV-Indian Institute of Information Technology and Management}
  \city{Gwalior}
  \state{Madhya Pradesh}
  \country{India}
  \postcode{474010}
}
\email{saxenapriyanshasd@gmail.com}

\author{Raahat Gupta}
\affiliation{
  \institution{ABV-Indian Institute of Information Technology and Management}
  \city{Gwalior}
  \state{Madhya Pradesh}
  \country{India}
  \postcode{474010}
  }
\email{raahat.gupta.1998@gmail.com}

\author{Akshat Maheshwari}
\affiliation{
  \institution{ABV-Indian Institute of Information Technology and Management}
  \city{Gwalior}
  \state{Madhya Pradesh}
  \country{India}
  \postcode{474010}
  }
\email{aks3d76@gmail.com}

\author{Saumil Maheshwari}
\affiliation{
  \institution{ABV-Indian Institute of Information Technology and Management}
  \city{Gwalior}
  \state{Madhya Pradesh}
  \country{India}
  \postcode{474010}
  }
\email{saumilmaheshwari@yahoo.co.in}

\renewcommand{\shortauthors}{Saxena, et al.}

\begin{abstract}
  Semantic inpainting or image completion alludes to the task of inferring arbitrary large missing regions in images based on image semantics. Since the prediction of image pixels requires an indication of high-level context, this makes it significantly tougher than image completion, which is often more concerned with correcting data corruption and removing entire objects from the input image. On the other hand, image enhancement attempts to eliminate unwanted noise and blur from the image along with sustaining most of the image details. Efficient image completion and enhancement model should be able to recover the corrupted and masked regions in images and then refine the image further to increase the quality of the output image. Generative Adversarial Networks (GAN), have turned out to be helpful in picture completion tasks. In this chapter, we will discuss the underlying GAN architecture and how they can be used used for image completion tasks.
\end{abstract}

\maketitle

\section{Introduction to GAN}
\label{sec:1}

\textbf{Generative Adversarial Network (GAN)} is a type of deep neural network that is becoming increasingly popular in these days. They have immense applications in doing tasks which were once considered to be too complicated for computers to solve. This chapter first gives an introduction of GAN along with intuitive examples, then lays out a brief description of its applications followed by one such application in detail -- \textit{how to use GAN for the task of image completion.} 

After reading this chapter, readers will be able to appreciate the beauty of generative networks and apply then to real-scale applications. Let us first start with a brief overview of the short albeit fascinating history of GAN.

\subsection{History}
\label{subsec:1.1}

There are several tasks where computers have matched, or even surpassed, human performance. Computers have come a long way and even exceeded human intelligence in fields like face recognition, classification, or even games such as \textit{Go}. However, there are still many applications where computers have a long way to go -- the most prominent among them being chatbots or voice-powered assistants like Alexa, Google Home, etc. One can easily distinguish a conversation with a voice-assistant versus a human. Thus, we can say that computers have not entirely passed the Turing test yet.

One of the main reasons for this "gap" of intelligence is that historically A.I. algorithms were remarkable at \textit{learning} various complex patterns and structures among data, and using them to make a decision, such as a classification or decision the next move in \textit{Go}. However, computers were not so good in \textit{generating} new data -- something that humans have to learn from a very early stage. 

This "gap" was significantly reduced when Ian Goodfellow et al. published the paper on \textbf{Generative Adversarial Networks (GAN)} in 2014\cite{survey-1}. While we cannot say that GANs were the first algorithms to enable computers to \textit{generate} new data -- other algorithms have existed before them -- but GANs were able to do the task significantly better than others and at degrees that matched production levels.

So, we can say that GANs enable us to accomplish milestones of artificial intelligence that were considered near-impossible, even just before the time the paper was published. Abilities of GAN include generating fake images in a real-world environment, removing objects from images, converting a video clip of a horse into a zebra, and many more! GANs are so remarkable that even industry experts like Yann LeCun, Director of AI research at Facebook, were quoted as saying that GANs and their variations are "the coolest idea in deep learning in the last twenty years."\footnote{LeCun, Yann. "RL Seminar: The Next Frontier in AI: Unsupervised Learning"}

\subsection{But, what \textit{exactly} is a GAN?}
\label{subsec:1.2}

In a nutshell, \textbf{Generative Adversarial Networks} use a aggressive, game-like environment to train two neural networks. The networks "compete" with each other, and the end result is an image or a sequence of words that are indistinguishable from what would appear in the real world.

GAN comprises of two networks -- the \textbf{Generator} and the \textbf{Discriminator}. The job of the generator is to manufacture data (images or text) that is indistinguishable from the real examples in the training set. The discriminator, on the other hand, is tasked with identifying the data coming from the generator versus (the \textit{fake} examples) the data in training set (the \textit{real} data).

The generator, on its way to produce realistic-looking images, receives feedback from the discriminator. The goal of the generator is to fool the discriminator as many times as possible. This gives it a metric to measure its performance, and this metric can effectively be used to train the generator.

The discriminator trains as well -- it is given feedback based on the percentage of images correctly classified as fake versus the fake images that get away. Its goal is to classify as much fake images as possible; so in a sense we can say that \textit{the generator and discriminator compete with each other} when a GAN is training. Both networks continue to improve as this cat-and-mouse game progresses.

One of the curious properties of GAN -- and one which makes them particularly hard to train -- is that the optimisation minimum is not fixed in GAN. Normally, gradient descent contains an optimisation function which tires to minimise a certain \textit{cost function}. However, in case of GAN, the optimisation function is seeking a balance between the two opposing forces; it continues to work until a state of \textit{equilibrium} is achieved. 

\subsection{An Intuitive Example}
\label{subsec:1.3}

Let us picture an analogy to understand the inner workings of GAN better. Consider a small town which has an organised crime unit -- the local mafia. The mafia tries to counterfeit money, and every time it produces a new batch of bills, it attempts to deposit them in the bank (via an associate, of course). 

If the associate gets arrested, the mafia learns that the bank has judged their bills as counterfeit and knows that it has to improve in the production of bills. Just like a \textit{generator} whose examples are rejected, the mafia does not know exactly where it went wrong; just that its attempt to fool the bank was not fruitful. It goes back to work, and its team of chemists and analysts produce another set of bills, probably better than the previous ones.

Similarly, the bank gets better at distinguishing counterfeit bills over time. It acts as the \textit{discriminator} and provides \textit{feedback} to the generator at the end of each iterator (by arresting their associate or not). The bank may decide to invest in better money scanning technologies, hire experts, etc. so that no counterfeit money gets by.

In this manner, the generator and discriminator both get to improve, and we are left with a bunch of high-quality counterfeit money at the end!



\section{GANs in Action}
\label{sec:2}

Now that everything is probably clear about the basic workings of GAN, we look at the specific architecture and terms associated with GANs in this section. The pseudocode of GAN training algorithm along with visualisation, is also described. We end this section by giving various applications where GAN is used in real-life.

\subsection{Architecture of GAN}
\label{subsec:2.1}

\begin{figure}[ht]
\centering
\includegraphics[scale=0.50]{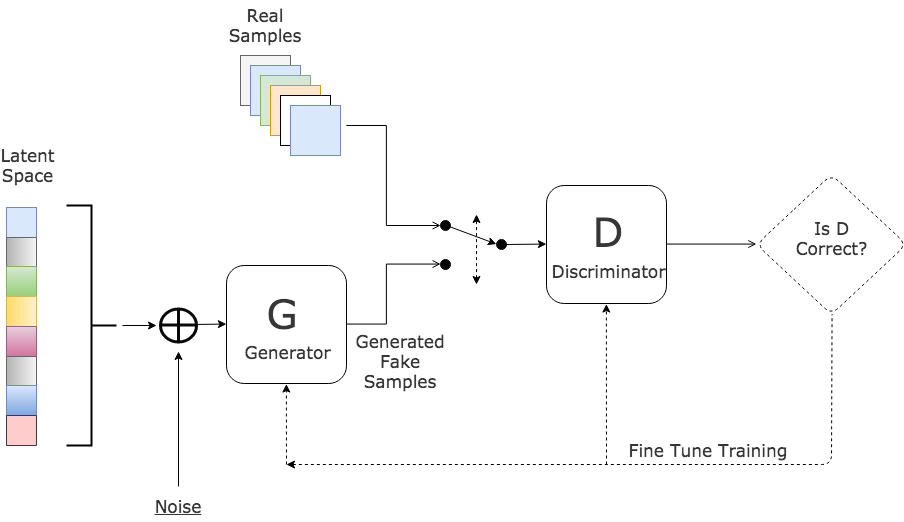}
\caption{The basic architecture of Generative Adversarial Networks\protect\footnotemark}
\label{fig:gan-basic-structure}
\end{figure}

Imagine our goal is to teach a GAN to produce realistic-looking handwritten digits. Figure \ref{fig:gan-basic-structure} shows the diagram of core GAN architecture.

\footnotetext{Source:  \url{https://www.linkedin.com/pulse/gans-one-hottest-topics-machine-learning-al-gharakhanian/?trk=pulse_spock-articles}}

The architecture can be classified into the following parts:

\begin{enumerate}
\item  \textbf{Training Dataset}: This is the set of real images shown in the upper left corner of the image. The real and labelled examples are shown only to the discriminator and not to the generator. Thus, the latter has to generate counterfeit images which seeing what an actual one looks like. The inputs are generally represented by $x$. 
\item \textbf{Random Noise Vector} : The input to the generator also contains a vector of random numbers ($z$) in addition to the latent space as shown in the left side of the image. The generator uses the random noise as a starting point of its synthesis process.
\item \textbf{Generator Network}: It is represented by $G$, it takes the random numbers $z$ and creates a set of fake examples $x*$. Its goal is so make these fake examples as indistinguishable from the real ones as possible.
\item \textbf{Discriminator Network}: The discriminator $D$ inputs both $x$ and $x*$, and gives a score (on a scale from 0 to 1) based on the identicality of these inputs. 
\item  \textbf{GAN Training}: For each of the discriminator's output, its performance is determined with respect to a predefined metric. This performance is used to update both generator and discriminator through backpropagation:
\begin{enumerate}
\item The Discriminator's weights and biases get updated to maximise its classification accuracy;
\item The Generator's weights and biases get updated to maximise the likelihood that the Discriminator incorrectly classifies $x*$ as real.
\end{enumerate}
\end{enumerate}

\subsection{GAN Training Algorithm}
\label{subsec:2.2}

As GAN consists of two entities Discriminator and Generator, the depiction of steps for both during training of GAN \cite{survey-2}, is given below:\\

\textbf{The Discriminator}
\begin{enumerate}
\item We take a random sample from the training dataset. Label it as $x$.
\item Use the generator network to produce a batch of fake images $x*$ (note this step includes taking a random noise vector $z$ first).
\item The discriminator is fed inputs $x$ and $x*$; it is used to classify among the two.
\item The classification errors and losses are computed, and, through the process of backpropagation, the weights and biases of the discriminator are updated accordingly. 

\textit{The goal of the discriminator is to minimize classification errors.} 
\end{enumerate}
 
\textbf{The Generator}
\begin{enumerate}
\item We take a new random noise vector $z$.
\item The generator uses $z$ as input to produce a batch of fake images $x*$.
\item This $x*$ is fed to the discriminator, which calculates the classification score based on $x$ and $x*$.
\item The error and loss are computed, and via backpropagation, the weights and biases of the generator are updated accordingly.

\textit{The goal of the generator is to maximize the classification errors.}
\end{enumerate}

The pseudocode of the GAN training is depicted in Algorithm \ref{alg:gan-a}.

\begin{algorithm}[H]
\caption{GAN Training Algorithm}
\label{alg:gan-a}
\begin{algorithmic}[1]
\State \textbf{for} each training iteration 
		\State \hspace*{0.5cm} Train the Discriminator
		\State \hspace*{1.0cm} Take a random sample from the training set. Label it as $x$.
		\State \hspace*{1.0cm} Take a new random noise vector. Label it as $z$.
		\State \hspace*{1.0cm} Utilizing the Generator, use $z$ to manufacture a counterfeit example $x*$ 
		\State \hspace*{1.0cm} Use the Discriminator and classify $x$ and $x*$. 
		\State \hspace*{1.0cm} Compute classification losses and use back-propagation for updating the discriminator weights and biases. Minimize classification errors.
		\State \hspace*{0.5cm} Train the Generator 
		\State \hspace*{1.0cm} Take a new random noise vector $z$. 
		\State \hspace*{1.0cm} Utilizing the Generator, use $z$ to manufacture a counterfeit example $x*$ 
		\State \hspace*{1.0cm} Use the Discriminator and label $x*$.
		\State \hspace*{1.0cm} Compute classification losses and use back-propagation for updating the generator weights and biases. Maximize Discriminator errors. 
\State end \textbf{for}

\end{algorithmic}
\end{algorithm}

\subsection{When to stop training in GAN?}
\label{subsec:2.3}

As stated earlier, the optimisation minimum is not fixed in the case of GAN. Our goal is to seek a balance between the two opposing forces -- the generator and the discriminator. This could create a state of ambiguity as to when to stop training. \textit{How can we be certain that the state of equilibrium is achieved, and further training will not necessarily benefit the GAN algorithm?}

This problem is somewhat similar to the \textit{zero-sum game} in Game Theory -- in this case one player's gains come from another's losses; and the exact value that one wins, the other loses. All zero-sum games include a point where neither player can improve their situation by changing any of their actions; such a point is called \textit{Nash Equilibrium}\footnote{Nash Equilibrium is named after the American economist and mathematician John Forbes Nash Jr, whose life and career were captured in the biography titled \textit{A Beautiful Mind} and inspired the eponymous film.}. Our goal is to find the Nash equilibrium in case of GAN models.

The generator and the discriminator reach their Nash equilibrium when the fake examples produced by the former are indistinguishable from the real data, and the latter can, at best, randomly guess whether a particular example is real or fake. Let's look at the following two cases:

\begin{itemize}
\item Although it may seem that the discriminator is just guessing real and fake at random 50\% probability (and hence has room to improve), but if the fake examples $x*$ are truly indistinguishable from the real examples $x$, then there is nothing the discriminator can do to tell them apart from one another. Thus, it cannot do better than a random chance probability.

\item The Generator is likewise at a point where it has nothing to gain from further tuning. Because the examples it produces are already perfectly indistinguishable from the real ones, even a tiny change to the process it uses to turn the random noise vector $z$ into a fake example $x*$ may give the discriminator a clue about how to tell apart the fake example from the real data, making the generator worse off.
\end{itemize}

A GAN is fully trained when Nash equilibrium is achieved. But, as if often the case with such optimisation problems, such an equilibrium is \textit{very} hard to reach in practice. We stop as soon as we achieve our desired objective at some judgement level. 

\subsection{Applications of GAN}
\label{subsec:2.4}

Although it is almost impossible to list the entire applications of GAN in this section, a few important and interesting ones are listed here:

\begin{itemize}
\item Generating photo-realistic fake images: Take a look at Figure \ref{fig:gan-c}. All the faces look real, aren't they? But the fact is none of the faces are real; they're all produced using \textit{Progressive Growing} of GAN \cite{survey-3}.  

\begin{figure}[h]
\centering
\includegraphics[scale=0.50]{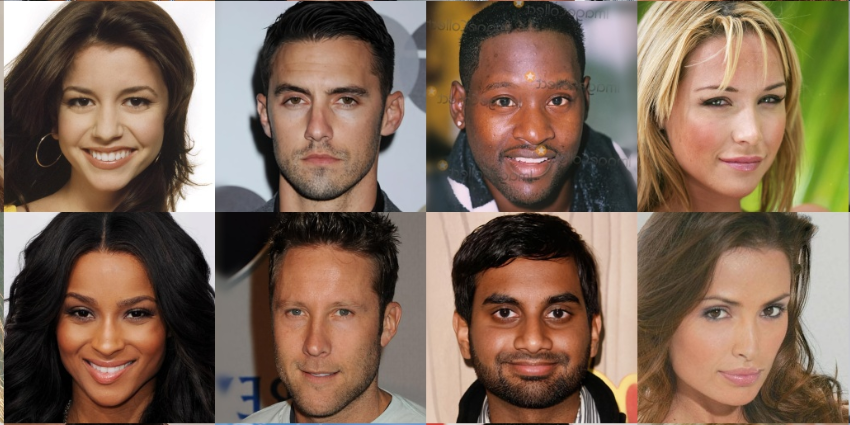}
\caption{Generating fake faces using GAN \cite{survey-3}}
\label{fig:gan-c}
\end{figure}

\item Image-to-Image Translation: We can have a little fun and use an implementation called \textit{CycleGAN} to replace a horse in an image with a zebra (and vice-versa) while keeping all other factors same (as shown in Figure \ref{fig:gan-d}). We can also do something more meaningful and convert a photograph into a monet \cite{survey-4}.

\begin{figure}[h]
\centering
\includegraphics[scale=0.50]{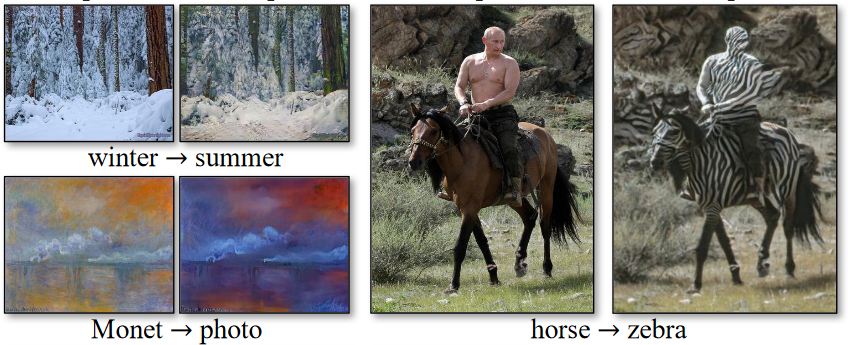}
\caption{Converting horses to zebras and monets to photos using GAN \cite{survey-4}}
\label{fig:gan-d}
\end{figure}

\item Automatic synthesis of realistic images from a textual sentence using \textit{Stack GAN}, and transferring style from one domain to another domain using \textit{Discovery GAN} \cite{survey-5}.

\item Generating realistic a image from attributes: Imagine that burglar breaks in your house at night. You catch a brief glimpse of him/her, but nothing useful enough to identify the person. Now, suppose there's an advance system in the police station that could  generate a realistic image of the thief based on the description provided by you. That system would probably use GAN \cite{survey-6}.
\end{itemize}


\section{Image Completion}
\label{sec:3}

This section will discuss about one of the most common and highly interesting application when Generative Adversarial Networks are used: the task of \textbf{Image Completion}. 

Image completion, or more formally, \textit{semantic inpainting} defines the task of completing arbitrary-sized unknown subsets of images based on semantics \cite{survey-7}. Our task is primarily to predict accurately high-level useful content; this is considerably difficult than just filling in the images with spurious data. Semantic implantation finds applications in restoration of damaged artworks, or editing images to remove inanimate objects from them, etc. An interesting range of applications and results are shown in sections below. 

\subsection{Related works}
\noindent Jia-Bin Huang and Ahuja have proposed an advance-knowledge approach which used contextual information for image completion. However, in case the corrupted region is large or is irrelevant to visual data, or if the complexity of the image is high, the output of the method would be quite unsatisfactory \cite{dummy:6}. Connelly Barnes and Eli have proposed patch matching algorithm for image completion for nonparametric texture construction. The algorithm performed satisfactorily and was able to identify similar patches. However, it failed when the original image lacked adequate data to complete the missing regions \cite{dummy:1}. Yunjin Chen and Thomas Pock have proposed nonlinear response dispersion model, which consists of a feed forward network with a fixed number of gradient descent stages. Trainable nonlinear reaction-diffusion accomplished promising execution in any case; its display was prepared for a specific noise level. It was unfit to perform well on pictures with obscure noise levels. Additionally, it requires the output which is expected by the network during training \cite{dummy:2}. In 2011, Deng transformed the inpainting task to the graph-labeling task using graph Laplace method. However, this method required images samples of the image to be inpainted be included in the training data, which was not practical in real life applications \cite{dummy:4}. A viable face inpainting algorithm utilizing a generative model was proposed by Yijun Li, Sifei Liu, and Jimei Yang. From background inpainting task, face inpainting is a challenging task because it regularly needs to produce semantically newer pixels areas in the missing region parts like eyes and nose, which can vary from person to person. Even though the model had the capacity to produce semantically conceivable and outwardly satisfying content, it has a few constraints. The model still could not deal with some unaligned faces. also, it did not wholly misuse the spatial connections between nearby pixels \cite{dummy:8}. Ruijun and Yang proposed an improved generative translation model. The paper proposed a semantic image completion method using regional completions for painting completion. Using the generator and discriminator network, the missing region is generated, which should be consistent with the surrounding region. However, image completion work is restricted to only face data and needed to be improved to ensure that the entire painting work could be recovered \cite{dummy:9}.\\ 

Deepak Pathak put forward Context Encoders(CE) which estimated missing areas in images based on its surroundings. However, during training it needed a mask on the corrupted regions of the image, that is a significant disadvantage of the approach, and also context encoders led to blurry and noisy results in the inpainted parts \cite{dummy:12}. Ren proposed a novel CNN architecture named Shepard Convolutional Neural Networks which efficiently equips conventional CNN with the ability to learn missing data. However, in case the corrupted region was large or was irrelevant to visual data, or if the complexity of the image is high, the output of the method would be quite unsatisfactory \cite{dummy:14}.\\

In \cite{dummy:15}, low-light enhancement model using convolutional neural network and Retinex theory was proposed. It showed an equivalence between multi-scale Retinex and feedforward convolutional neural network using Gaussian kernels. However, because of the limited receptive field in their model, very smooth regions such as clear sky are sometimes attacked by the halo effect. Jeremias sulam, formulated trainlets, to construct large adaptable atoms using various datasets of facial images using dictionary learning algorithm. Because of the computational constraints, this method was applied to tiny regions of the image and not on the entire image. As a result, this approach did not give satisfactory results on large regions in images \cite{dummy:17}. Raymond and Chen \cite{dummy:19} proposed another picture completion technique that can be utilized to fix any state of gaps. In any case, such training depends on the data used in training. In the meantime, the processing of surface and structure was not sufficiently impeccable. Kai Zhang and Yunjin Chen proposed a picture denoising approach in which they built feed-forward denoising convolutional neural systems using residual learning and batch normalization. However, this methodology was unfit to recover missing regions, and it just denoised the picture. Likewise, it was unfit to refine pictures with genuine complex commotion and other general picture restoration tasks \cite{dummy:20}.\\ 
 

\section{Introducing Wasserstein GAN for Image Completion}
\label{sec:4}

Now that you've known what image completion really is, let us look at how it can be solved with GANs. Though other GAN architectures are available, we are using the one called \textbf{Wasserstein GAN} for this purpose. Wasserstein GAN is an architecture that can be used for image completion tasks. It creates the coarse patches to fill the missing region in the distorted picture, and the enhancement network will additionally refine the resultant pictures utilising residual learning procedures and hence give better complete pictures for computer vision applications. The algorithms is described in the following sections. For an overview of the results in the CelebA-hq dataset, you can go directly to Section \ref{sec:5}.

\subsection{Methodology}
\label{subsec:4.1}

The methodology could be separated into three different steps. 

In the first step, data-preprocessing on CelebA-hq dataset\footnote{\url{http://mmlab.ie.cuhk.edu.hk/projects/CelebA.html}} is done to train and test the developed model. 

The following data preprocessing steps were followed:
\begin{itemize}
\item The dataset is splited into 15000 training images and around 1000 testing images.
\item Each face image in the dataset is resized to 64* 64* 3 pixels to train the Wasserstein GAN model.
\item Masking- A binary mask is used with values 0 or 1. 0 corresponds to the corrupted region while 1 corresponds to the uncorrupted region in the image. This binary mask is applied to all images  to make them corrupted which will serve as input of the training process.
\end{itemize}

In the second step, a Wasserstein GAN based model to complete the missing pixels in the image is developed. The image completion GAN gives a complete image with a blurry filled area. The generator of the GAN generates real looking images, but in the process of generation, the noise gets unavoidably added.

So, in the third step, the output of the generator is passed through the enhancement network to make the filled area clear and to refine the completed image further. The enhancement network is trained using 2000 image pairs containing blurry images and its corresponding clean images.

\subsection{Wasserstein Distance as Loss}
\label{subsec:4.2}

This architecture is different from the one proposed by Goodfellow\cite{survey-1} in that it uses Wasserstein distance as to train the generator so that it can capture training data distribution and generate images similar to those in the training data.

Wasserstein distance is a measure of the distance between two probability distributions. For the generated data distribution $p_g$ and the real data distribution $p_r$, it can be mathematically defined as the cost for the cheapest plan from $p_g$ to $p_r$. It is also called Critic loss or Wasserstein distance. The Wasserstein distance loss function $L$ to train the generator can be mathematically represented as:

\begin{equation}
L = \mathbb{E}_{\tilde{x} \sim p_g} |C(\tilde{x})| - \mathbb{E}_{x - p_r} |C(x)|
\end{equation}

Here, the first term represents the expectation of the distribution generated by the generator, and the second term represents the expectation of the real training data distribution. By minimising the difference between the two, the generator learns to generate samples having probability distribution similar to training data distribution. Now, to make the learning faster and make model convergence faster gradient penalty term is added to our loss function. So, the overall loss function $L$ of Wasserstein GAN becomes: 

\begin{equation}
L = \mathbb{E}_{\tilde{x} \sim p_g} |C(\tilde{x})| - \mathbb{E}_{x - p_r} |C(x)| + gradient  penalty
\end{equation}

where, gradient\,penalty will be given by
\begin{equation}
gradient\,penalty=\lambda\displaystyle\mathop{\mathbb{E}}_{\hat{x}\sim P_{g}}[(||\bigtriangledown_{\hat{x}} C(\hat{x})||_{2}-1)^2]   
\end{equation}
here $\lambda$ is the  gradient penalty coefficient.

\subsection{Image Generation using Wasserstein GAN}
\label{subsec:4.3}

After training the generator to generate samples which look real, the next aim is to ensure that the missing region generated has a similar context to the non-missing region so that sensible looking completed images as output can be obtained.

A binary mask with values 0 or 1 is used. 0 corresponds to the corrupted region while 1 corresponds to the uncorrupted region in the image. Let $y$ represents the uncorrupted image. $M \odot y$ gives the uncorrupted part of the image. Let $G(z')$ be some image generated by the generator which suitably completes the missing region in the image. $(1-M) \odot G(z')$ represents the completed region which when added to the uncorrupted region gives the reconstructed image as output\cite{survey-8}:

\begin{equation}
x_{reconstructed} = M \odot y + (1 - M) \odot G(z')
\end{equation}

To find $z'$ that suitably completes the image following loss functions are defined:

\textbf{Contextual Loss: } To ensure both generated and the input image have same context, ensure that the uncorrupted pixel  in original image y are same as the pixels in the generated image G(z) at a particular location. For this, pixel wise difference between the uncorrupted part of the two images is taken and then this difference is minimized.
\begin{equation}
     L_{contextual}(z) = ||M\odot G(z) - M\odot y ||_1 
\end{equation}
where $||x||_1$ represents $l_1$ norm of some vector x. \\

\textbf{Perceptual Loss: } It ensures that the output image looks real. For this, the following perceptual loss:

\begin{equation}
L_{perceptual}(z) = log(1 - C(G(z)))
\end{equation}

\textbf{Total Loss: } It is a sum of perceptual and contextual loss and is denoted by L(z):
\begin{equation}
    L(z) = L_{contextual}(z) + Q L_{perceptual}(z)
\end{equation}
\noindent Q is a hyper-parameter and we minimize this loss function to ensure completed image is contextally similar to input image.

\subsection{Enhancement Network}
\label{subsec:4.4}

\begin{figure}[h!]
	\begin{center}
		\includegraphics[scale=0.25]{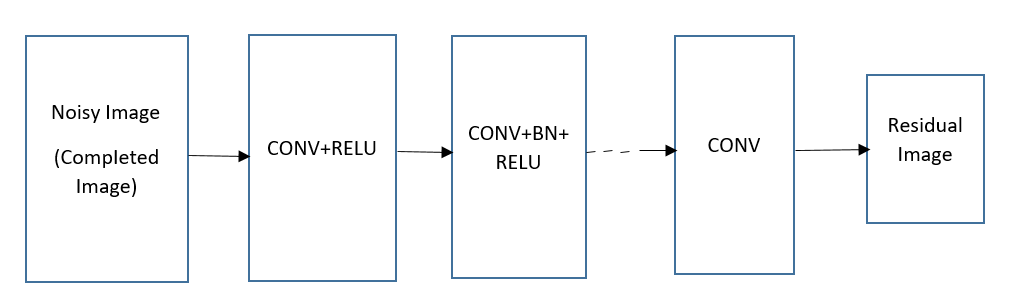}
		\caption{Enhancement network to refine completed images}
	\end{center}
\end{figure}
In enhancement network to refine completed images, the residual learning approach is used. The input to the network is blurry image y = x + v, here x is the clear image,v represents the blur added. The residual network is trained to grasp the mapping R(y)\( \approx \)v , to get the clear image x as x = y- R(y). 
Mathematically, the average mean square error among the output residual image by the model and the actual residual images is used as error function for getting the parameter \( \Theta \) to train the enhancement network. 
\begin{equation}
L(\theta)=\frac{1}{2N}\sum_{i=1}^{N}||R(y_{i};\theta)-(y_{i}-x_{i})||^2 
\end{equation}
Here, L is the training error of the enhancement network and N are total training images. Enhancement network consists of following layers as shown in Fig 4:
(i) Conv+ReLU: It creates feature maps, and ReLU adds the non-linearity.
(ii) Conv+BN+ReLU: This layers contains added batch normalization between Conv and ReLU.
(iii) Conv: It is used to get the output residual image. 

\subsection{Results and discussion}

The following plot was obtained by training the enhancement network on 2000 celeba-hq image pairs of clean and its corresponding blurr images.

\begin{figure}[h!]
	\begin{center}
		\includegraphics[scale=0.55]{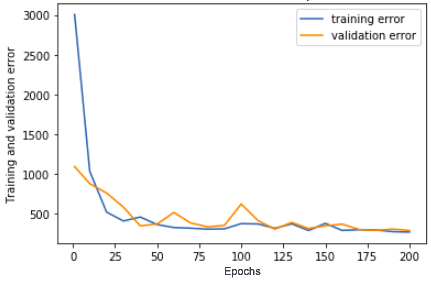}
		\caption{Enhancement network training plot to refine completed images}
	\end{center}
\end{figure}

In Fig 5 we observe that, as the training proceeds, the average mean square error among the output residual image by the model and the actual residual images decreases. As a result, according to Eq (8), the training error decreases. Finally, around 200th epoch, the enhancement network is sufficiently trained ,which is evident as the training error becomes constant at a particular value, and there is no further decrease.

The Wasserstein GAN model is trained on 15000 Celeba-hq images for 10000 epochs and batch size of 128. It is seen that in the initial stages of learning the expectation of the distribution generated by the generator is different from the expectation of the distribution of real data and hence the difference between the two is higher resulting in higher Wasserstein distance values. However, as the learning proceeds generator learns the distribution of the real data and then generates samples having a similar distribution with the real data, and hence the difference in their expectation decreases resulting in lower Wasserstein distance values. Now, around 10000 epochs the generator has sufficiently learned, and hence the Wasserstein distance values do not decrease further and becomes constant around a particular lower value. \\

In the initial stages, the context in the uncorrupted region of the generated samples and the original samples are different, so from Eq (5), it can be seen that the resulting contextual loss is higher. As the training moves further using Adam's optimizer (z) gets trained, and hence, there is a significant decrease in the contextual loss values. Around 1200$^{th}$ epoch, the context in the uncorrupted region of the generated samples and the original samples becomes quite familiar, and hence the contextual loss becomes constant around a particular value. 

Initially the distribution of generated images and real images is different, so the critic is able to distinguish the generated samples from the real ones and hence the value of C(G(z)) is close to 0 and as a result 1-C(G(z)) becomes close to 1 as a result from Eq (6) the loss is higher. However, as learning proceeds, the generator generates real looking samples as a result C(G(z)) becomes close to 1 and 1-C(G(z)) becomes close to 0, resulting in lower perceptual loss values from Eq (6). 

\begin{figure}[hbt!]
	\begin{center}
		\includegraphics[scale=0.55]{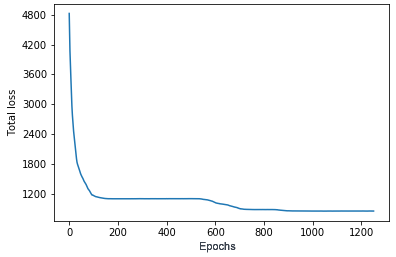}
		\caption{Total image completion loss plot}
	\end{center}
\end{figure}

The perceptual loss values are quite lower compared to contextual loss values, and as a result from Eq (7), the total image completion loss is almost equal to contextual loss, and as a result Fig 7 which is total image completion loss plot is almost similar to contextual loss plot for image completion.

The following two evaluation metrics to evaluate the quality of the output images by the model:

\subsection{Peak Signal-to-Noise Ratio (PSNR):}\label{AA}

PSNR \cite{dummy:5} is measured in decibels (dB). The higher the PSNR, the better image has been completed to match the original image.
\begin{equation}
MSE=\frac{1}{mn}\sum_{i=0}^{m-1}\sum_{j=0}^{n-1}||f(i,j)-g(i,j)||^2 
\end{equation}
\begin{equation}
PSNR=20\log_{10}(\frac{MAX_{f}}{\sqrt{MSE}}) 
\end{equation}
\noindent Here,
f is the original image,
g represents completed image through the model,
m represents image pixel rows,
n represents image pixel columns, i and j represents row and column index respectively.
MAX$_{f}$ is a constant equal to 255.

\section{{ Structural Similarity Index (SSIM):}}

\noindent The Structural Similarity (SSIM) Index \cite{dummy:5} depends on computation of terms, namely the luminance, contrast and structural term.
\begin{equation}
SSIM(x,y)=[C(x,y)]^\alpha\times[I(x,y)]^\beta\times[S(x,y)]^\gamma
\end{equation}

\noindent where, C represents contrast, I represents luminance, S represents structural term, x represents original, y represents completed images. The parameters $\alpha > 0$, $\beta > 0 $, and $\gamma > 0$, are used to adjust the relative importance of the three components.

The following \textit{PSNR} and \textit{SSIM} values through the proposed approach are compared with the existing techniques in Table 1:

\begin{table}[ht]
\centering
\caption{Comparison of PSNR values}
\begin{tabular}{ p{1.5cm} p{1.3cm}p{1.3cm} p{3cm}}
\hline\noalign{\smallskip}
 	\textbf{Methods} & \textbf{CE\cite{dummy:12}}  &\textbf{PI \cite{dummy:3}} & \textbf{Proposed approach}\\
 	\noalign{\smallskip}\hline\noalign{\smallskip}

	\textbf{PSNR(dB)} &	22.85  & 21.45  &  23.41\\
	\textbf{SSIM} &	0.872  &  0.851  &  0.9074\\

\end{tabular}
\end{table}

\noindent It can be seen that the approach performs well CelebA-hq dataset compared to other proposed image completion techniques, which is evident from the above PSNR and SSIM values.

Some of the results obtained through the proposed image completion approach using Wasserstein GAN are shown below in Figure 7:

\begin{figure}[ht]
\hspace{0.0in} Original \hspace{0.6in} Input \hspace{0.6in} Output \\

\centering
\subfloat{\includegraphics[width = 0.60 in]{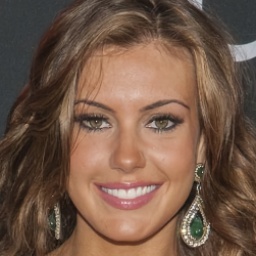}} \hspace{7.5 mm}
\setcounter{subfigure}{0}
\subfloat{\includegraphics[width = 0.60 in]{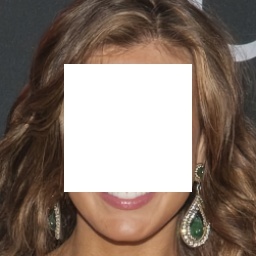}} \hspace{7.5 mm}
\subfloat{\includegraphics[width = 0.60 in]{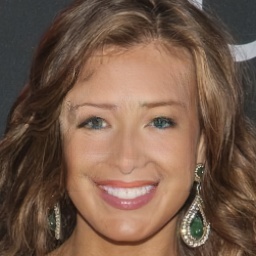}} \\ 
\subfloat{\includegraphics[width = 0.60 in]{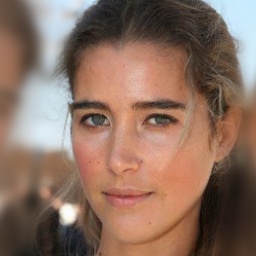}} \hspace{7.5 mm}
\setcounter{subfigure}{1}
\subfloat{\includegraphics[width = 0.60 in]{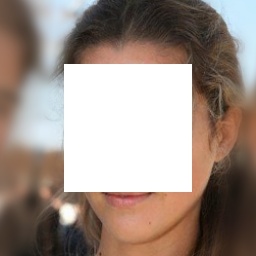}} \hspace{7.5 mm}
\subfloat{\includegraphics[width = 0.60 in]{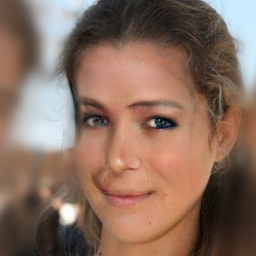}} \\ 
\subfloat{\includegraphics[width = 0.60 in]{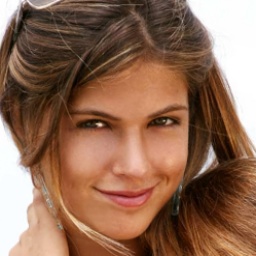}} \hspace{7.5 mm}
\setcounter{subfigure}{2}
\subfloat{\includegraphics[width = 0.60 in]{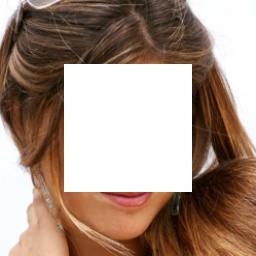}} \hspace{7.5 mm}
\subfloat{\includegraphics[width = 0.60 in]{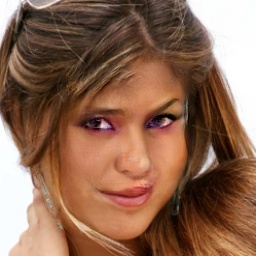}} \\ 
\subfloat{\includegraphics[width = 0.60 in]{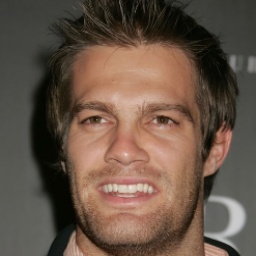}} \hspace{7.5 mm}
\setcounter{subfigure}{3}
\subfloat{\includegraphics[width = 0.60 in]{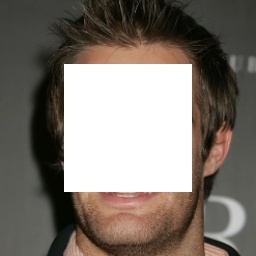}} \hspace{7.5 mm}
\subfloat{\includegraphics[width = 0.60 in]{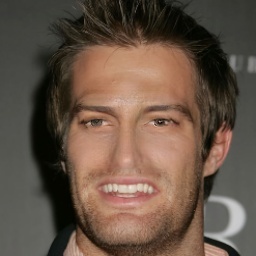}} \\ 
\subfloat{\includegraphics[width = 0.60 in]{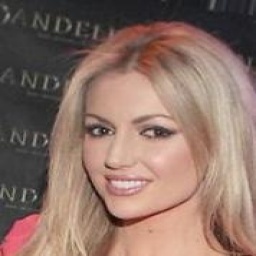}} \hspace{7.5 mm}
\setcounter{subfigure}{4}
\subfloat{\includegraphics[width = 0.60 in]{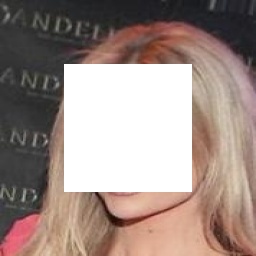}} \hspace{7.5 mm}
\subfloat{\includegraphics[width = 0.60 in]{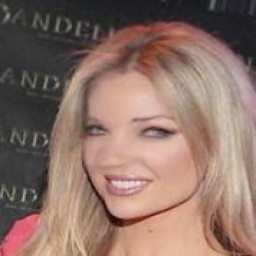}} \\ 
\caption{Experimental Results}
\label{some example}
\end{figure}


\section{Applications of Image Completion using GAN}
\label{sec:5}

\begin{figure}[h]
\centering
\includegraphics[scale=0.60]{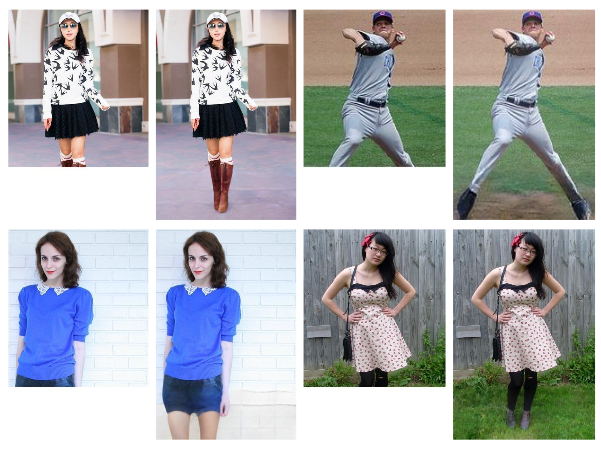}
\caption{Portrait Completion using GAN \cite{survey-9}}
\label{fig:gan-ex2}
\end{figure}

\begin{figure}[h]
\centering
\includegraphics[scale=0.45]{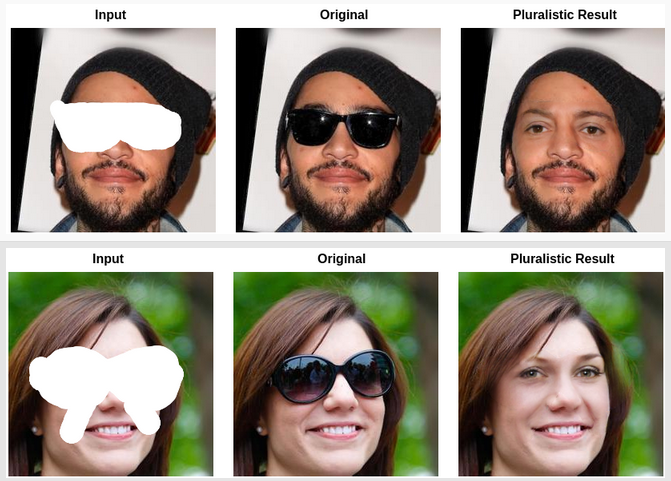}
\caption{Sunglasses Removal using GAN\protect\footnotemark}
\label{fig:gan-ex3}
\end{figure}

\begin{figure}[h]
\centering
\includegraphics[scale=0.45]{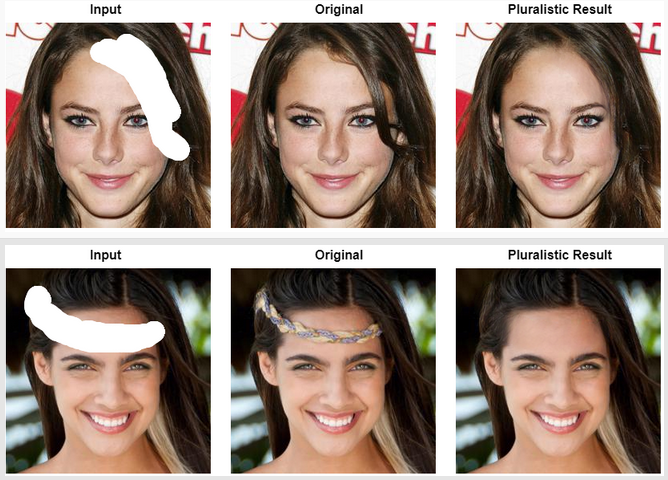}
\caption{Headband Removal using GAN\protect\footnotemark}
\label{fig:gan-ex4}
\end{figure}

This section looks at some of the real-world applications where GANs are useful.As depicted by the results in Figures \ref{fig:gan-ex2} -- \ref{fig:gan-ex4}, GAN does an astonishing job and the generator images are mostly alike to the actual/probable images.

\textbf{Portrait Completion}
Sometimes parts of a portrayal are missing from the image, due to the image being cropped, or an opaque article blocking our object of interest, etc. GANs are used in this scenario to complete the portrait as if the blocking article was never there \cite{survey-9}. Figure \ref{fig:gan-ex2} depicts the results obtained.

\textbf{Sunglasses Removal}
Not only are sunglasses worn to look more attractive, they can also hide the identity of lawbreakers and suspects. Thus, sunglasses removal is an active problem used by law enforcement to identify sunglasses-wearing suspects in surveillance footage with their known photographs. Results of the same are depicted in Figure \ref{fig:gan-ex3}.

\footnotetext{Source: \url{http://www.chuanxiaz.com/project/pluralistic/}}

\textbf{Object Removal on Face}
Similar to the above problem, it also involves removing an inanimate or unwanted object on the face. Examples include removing that headband that went out of fashion years ago, or removing a ring from one's finger etc. Figure \ref{fig:gan-ex4} shows impressive results in this domain.

\footnotetext{Source: \url{http://www.chuanxiaz.com/project/pluralistic/}}

%
%

\section{Summary}
\label{sec:6}

This chapter introduced GANs and their working. The intuition behind the working of GAN and the relevant mathematics of the topic were also provided. To summarize, GANs are made up of the following unit:

\begin{itemize}
\item \textbf{The Generator:} with the goal of fooling the discriminator by manufacturing instances that are indistinguishable from the training dataset; and

\item \textbf{The Discriminator:} with the goal of correctly identifying which instances are taken from the legitimate training set, and which are manufactured by the generator.
\end{itemize}

GANs can be used to create hyperrealistic imagery, i.e., imagery that is completely artificially generated by computers but looks completely real to the human eyes. On a philosophical note, all technological innovations have misuses. And since it is seemingly impossible to uninvent a technology, it is the job of scholars and researchers to keep this technology safe and work actively towards achieving its substantial potential \cite{survey-2}.

In this chapter, we were only able to scratch the surface of what is possible with GANs; however, we hope that after reading this chapter you have the necessary basic theoretical knowledge to continue exploring any facet of this field that you find most interesting.

%
%

\end{document}